\pdfoutput=1

\documentclass[11pt]{article}

\usepackage[final]{acl}

\usepackage{times}
\usepackage{latexsym}

\usepackage[T1]{fontenc}

\usepackage[utf8]{inputenc}

\usepackage{microtype}

\usepackage{inconsolata}

\usepackage{graphicx}

\usepackage[capitalize]{cleveref}
\crefname{section}{Sec.}{Secs.}
\Crefname{section}{Section}{Sections}
\crefname{table}{Tab.}{Tabs.}
\Crefname{table}{Table}{Tables}
\usepackage{color, colortbl}
\definecolor{Gray}{gray}{0.93}
\usepackage{multirow}
\usepackage{svg}
\usepackage{amssymb}

\usepackage{threeparttable}
\usepackage{multirow}
\usepackage{booktabs}
\usepackage{adjustbox}

\title{SHuBERT: Self-Supervised Sign Language 
Representation Learning\\via Multi-Stream Cluster Prediction}

\author{Shester Gueuwou$^{1}$, Xiaodan Du$^{1}$, Greg Shakhnarovich$^{1}$, Karen Livescu$^{1}$, Alexander H. Liu$^{2}$ \\ 
\small $^{1}$TTI-Chicago, $^{2}$MIT CSAIL \\  
\small{\{shesterg,xdu,greg,klivescu\}@ttic.edu, alexhliu@mit.edu}\\
\small{\url{http://shubert.pals.ttic.edu}}
}

\begin{document}
\maketitle

\begin{abstract}
Sign language processing has traditionally relied on task-specific models, limiting the potential for transfer learning across tasks. Pre-training methods for sign language have typically focused on either supervised pre-training, which cannot take advantage of unlabeled data, or context-independent (frame or video segment) representations, which ignore the effects of relationships across time in sign language. We introduce SHuBERT (Sign Hidden-Unit BERT), a self-supervised contextual representation model learned from approximately 1,000 hours of  American Sign Language video. SHuBERT adapts masked token prediction objectives to multi-stream visual sign language input, learning to predict multiple targets corresponding to clustered hand, face, and body pose streams. SHuBERT achieves state-of-the-art performance across multiple tasks including sign language translation, isolated sign language recognition, and fingerspelling detection. 

\end{abstract}
  
\section{Introduction}
\label{sec:intro}

 Sign language presents unique challenges compared to other language modalities, because of the relative scarcity of data and its multi-channel nature, combining manual, facial, and other body movements, which can be quick and highly coarticulated~\cite{bellugi1972comparison}.  Existing approaches to sign language processing have typically relied on models designed and trained for specific tasks, such as sign language translation (SLT) from signed to written languages~\cite{camgoz2018neural, shi2022open, zhang2024scaling}, isolated sign language recognition (ISLR)~\cite{kezar2023sem}, and fingerspelling detection and recognition~\cite{shi2019fingerspelling, fayyazsanavi2024fingerspelling, georg2024fsboard}. 
Pre-training approaches allow for pooling data across tasks, and several pre-training methods have been successful for sign language tasks~\cite{uthus2023youtube, rust-etal-2024-towards}. 
 However, these have typically focused on either supervised pre-training, which cannot take advantage of unlabeled data, context-independent (frame or video segment) representations, which ignore the effects of relationships across time in sign language, or contextual representations of only some aspects of sign language (see~\cref{sec:related_work}).
These limitations have historically constrained the performance and scalability of sign language processing systems.

\begin{figure}[t!]
  \centering
  \includegraphics[width=\linewidth]{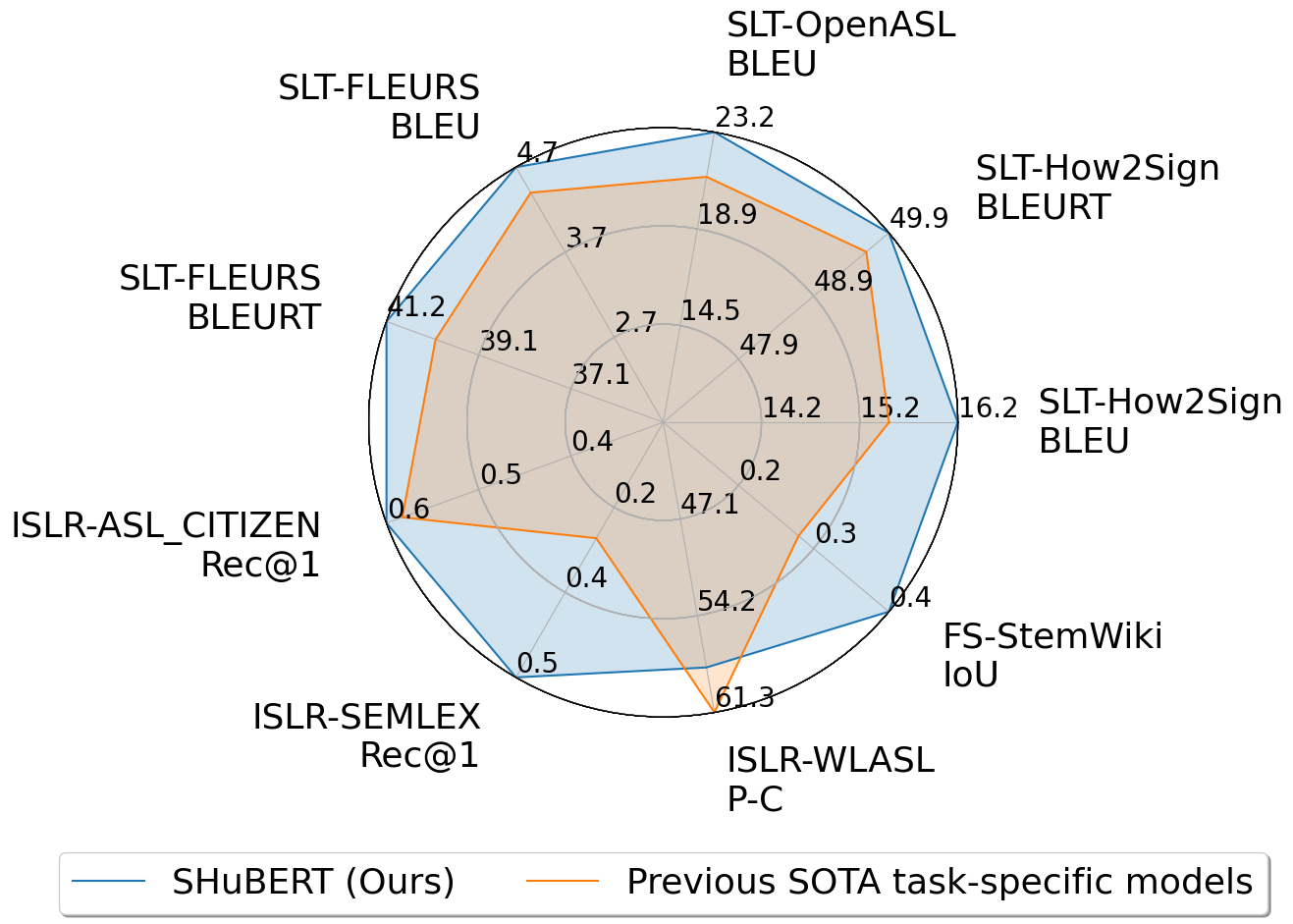}
  \caption{Comparison between our results using fine-tuned SHuBERT and results of the  previous state-of-the-art task-specific models on a suite of tasks, datasets, and metrics. Note: The orange shade does not represent a single model but a collection of the previous SOTA results for models trained on public data.
  The SHuBERT-based results improve on all but one task-specific SOTA model. See \cref{sec:setup} for details.
}
\vspace{-10pt}
\label{fig:radar}
\end{figure}

\begin{figure*}[t!]
    \centering
    \includegraphics[width=\linewidth]{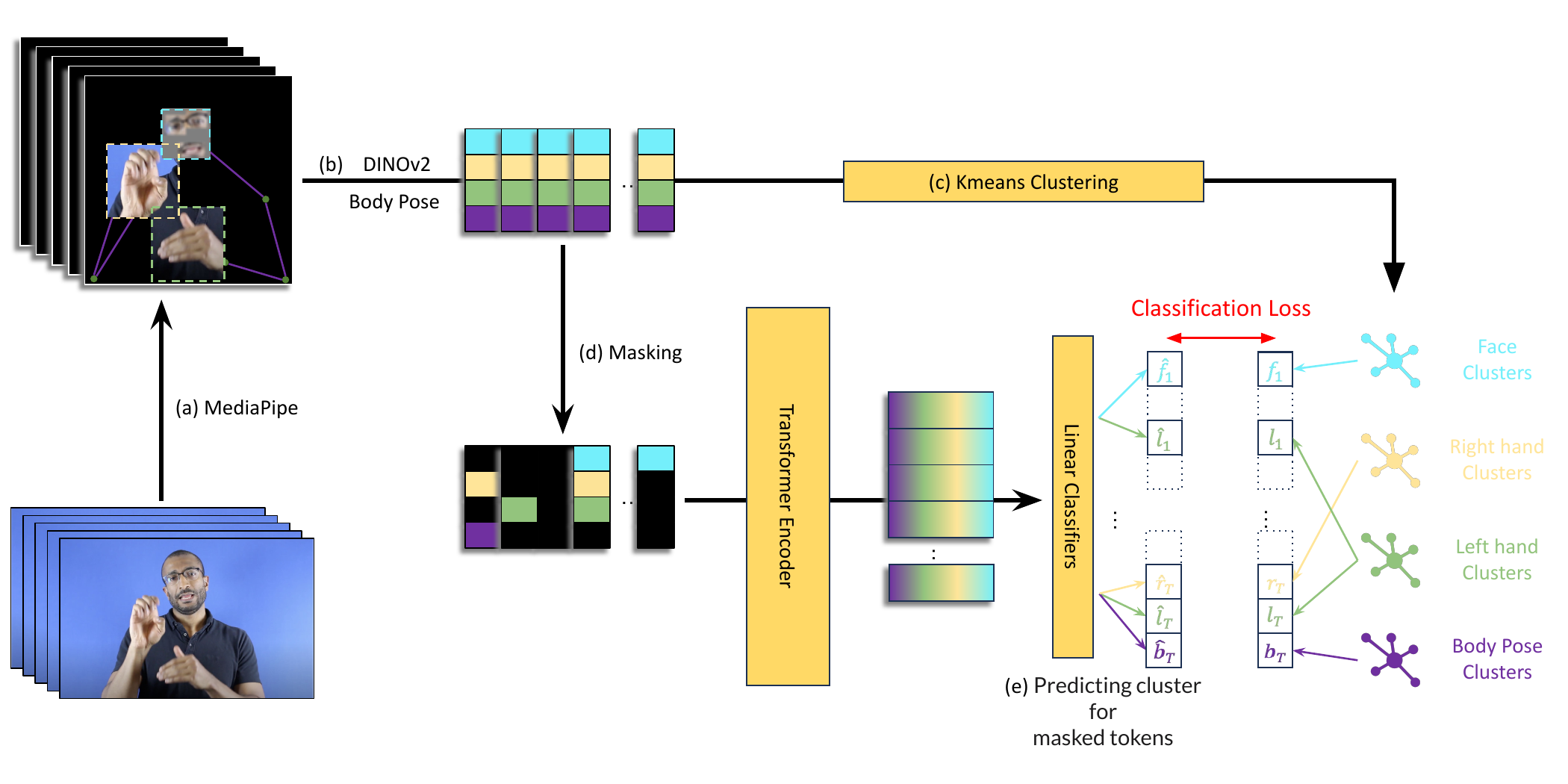}
    \caption{
SHuBERT pre-training. (a) We locate a set of landmarks in each frame of the input video using MediaPipe~\cite{lugaresi2019mediapipe},
with inter-frame interpolation to fill in missing landmarks. From these, we extract the upper body pose, %
crop the hand and face regions, and blur and partially mask the face crop %
for a measure of privacy and robustness. (b) We use DINOv2~\cite{oquab2024dinov} to extract features for the hands and face, 
yielding a four-stream representation (two hands, face, body pose) for each frame. (c) We assign the feature vectors for frame $t$ to cluster indices using pre-computed $k$-means clusters, yielding assignments $(f_t,l_t,r_t,b_t)\in [k]^4$ for face, left and right hand, and body pose, respectively. (d) We partially mask the features, and the masked features form the input to the transformer encoder. (e) We train SHuBERT to predict the cluster assignments for each masked input frame, $(\widehat{f}_t,\widehat{l}_t,\widehat{r}_t,\widehat{b}_t)$.}
    \label{fig:shubert_pipeline}
\end{figure*}

The success of self-supervised representations 
for written and spoken language, 
such as BERT for written language~\cite{devlin2019bert} and HuBERT for speech~\cite{hsu2021hubert}, has yet to be realized for sign language.
Self-supervised learning seems particularly relevant for sign language, for which annotated datasets are scarce. But the unique multi-channel and other visual properties of sign languages
suggest a specialized approach.

In this work, we present SHuBERT (Sign Hidden-Unit BERT) (\cref{fig:shubert_pipeline}), a self-supervised representation learning approach that learns \emph{contextual frame representations for all sign language channels jointly}. 
 SHuBERT adapts the masked prediction paradigm of BERT and HuBERT to the characteristics of sign language video, and learns %
 by predicting cluster assignments of multiple masked 
feature streams representing the hands, face, and body pose.  
The learned representations transfer effectively to multiple sign language understanding tasks, achieving state-of-the-art performance on several SLT benchmarks, multiple ISLR benchmarks, and fingerspelling detection, and improving over specialized models for each task (\cref{fig:radar}). %

\section{Related Work}
\label{sec:related_work}

Sign language understanding (recognition and translation) tasks have received increasing attention in the last few years~\cite{camgoz2018neural, shi2022open, lin2023gloss, kezar2023sem}.  
For translation, early work mainly focused on gloss-based methods, which rely on (the rare and small) datasets with manually labeled glosses~\cite{camgoz2018neural}. More recent work has turned to larger and more naturalistic datasets without gloss labels.
The most commonly used datasets are in American Sign Language (ASL)~\cite{duarte2021how2sign,shi2022open,uthus2023youtube}, German Sign Language (DGS)~\cite{camgoz2018neural}, British Sign Language~\cite{Albanie2021bobsl}, and Chinese Sign Language~\cite{zhou2021backtranslation}.  Of these, recent ASL datasets are the most naturalistic, and include %
large quantities of natively produced sign language (rather than translated from a spoken language, which has properties of ``translationese''~\cite{desai-etal-2024-systemic}).  For this reason we focus on ASL data and tasks, but our approach is applicable and extensible to any sign language.

\subsection{Pre-Training for Text And Speech} \label{sec:text_and_speech}

Pre-training is a cornerstone of modern language processing across modalities. 
For written language, encoder models like BERT~\cite{devlin2019bert} and its variants (e.g.,~\cite{liu2019roberta, lan2020albert}), based on masked language modeling, have served as dependable representations %
for language understanding tasks.
In speech processing, self-supervised learning approaches~\cite{mohamed2022self} have taken inspiration from text encoder models while addressing the unique challenges posed by continuous audio, which has no inherent segmentation into tokens nor a pre-defined token vocabulary.  For example, Hidden-Unit BERT (HuBERT) \cite{hsu2021hubert} adapts BERT by adding an offline clustering step to provide pseudo-labels for masked prediction. 
Such self-supervised representations, combined with task-specific fine-tuning, remain the state of the art for many speech tasks. 
 
 Sign languages share similar challenges to speech, with no pre-existing token lexicon and variable-length units (gestures) with no explicit boundaries, and our approach takes inspiration from HuBERT.  However, sign language video has its own unique challenges:
 the many sources of variation (signer appearance, background, lighting, camera angles), the multiple streams of gestures (hands, face, body), the high dimensionality of video, and the relative dearth of data.  These challenges are addressed in SHuBERT by focusing on the relevant streams (via pose tracking) combined with multi-stream masking and clustering.

\subsection{Pre-Training for Sign Language} %
\label{sec:related_sign}

\paragraph{Supervised pre-training.}
For translation of sign language video, the supervised pre-training approach has focused on (pre-)training a translation model on a large (but often noisy) out-of-domain dataset, followed by fine-tuning on a smaller in-domain dataset.  This type of pre-training leverages large collections of annotated data, with some systems~\cite{uthus2023youtube,tanzer2024youtubesl25,tanzer2024fingerspelling, zhang2024scaling}
trained on up to 6,600 hours of sign language content \cite{uthus2023youtube} to achieve state-of-the-art performance. However, much of this data remains private.
In addition, these approaches often involve substantial computational resources: The supervised model of ~\citet{zhang2024scaling}, for example, was trained on 128 TPU-v3 chips for 20 days. \citet{jiao2024visual} propose an alternative pre-training approach that greatly improves efficiency by using pose information only (rather than image pixels); however, this approach pre-trains and fine-tunes on the same training data.
Unlike these approaches, Uni-Sign~\cite{li2025unisign} is a supervised pre-training approach, based on mT5~\cite{xue2021mt5}, that has been applied to multiple tasks including both translation and ISLR.
Like all supervised methods, these approaches 
can not take advantage of available unlabelled sign language data.

\paragraph{Self-supervised pre-training.}
Previous work has compared multiple context-independent self-supervised techniques for ISLR, finding masked autoencoders (MAE) particularly effective~\cite{sandoval2023self}.
SSVP-SLT~\cite{rust-etal-2024-towards} adapts MAEs
 for large-scale sign language pre-training, achieving competitive performance on ASL-to-English translation.  This approach is computationally demanding, using 64 A100 GPUs for 14 days, and takes a maximum of 128 input frames ($\sim$8 seconds) at a time so is unable to model longer-term dependencies.  Other lines of work on SLT (e.g., \citet{chen2022simple}) have used a pre-trained S3D model~\cite{xie2018rethinking}, which requires the video sequence to be segmented into chunks, with each chunk treated as independent.
All of these approaches learn context-independent representations of individual frames or video segments, whereas SHuBERT learns contextual frame representations and can operate on long video directly. 

The only previous self-supervised approach of which we are aware for \emph{contextual} sign representation learning 
is SignBERT+~\cite{hu2023signbert+}, which extends the earlier SignBERT~\cite{hu2021signbert}.  This approach learns a representation specifically for hand poses,
via masked reconstruction of hand joints, 
and has strong results on ISLR, continuous (gloss-based) sign recognition, and sign translation on the RWTH-PhoenixT German Sign Language dataset~\cite{camgoz2018neural}.  However, this approach is inherently limited by not modeling the face and global body pose, and the results are obtained by combining SignBERT+ with a dataset-specific image pixel (RGB) representation.  In addition, SignBERT+ is pre-trained on the union of datasets on which it is tested; that is, it is exposed to the fine-tuning data during pre-training.  In contrast, SHuBERT models all components of sign language jointly, and is pre-trained on data that is disjoint from the fine-tuning data for the downstream tasks.

\subsection{Multi-Stream Models of Sign Language}
\label{sec:related_sign_multistream}

Several previous methods have taken advantage of the observation that sign language naturally decomposes into multiple streams of hand, face, and body motions.\footnote{The term ``multi-stream'' has been used in different senses in prior work.  For example, DSTA-SLR~\cite{hu2024dynamic} creates multiple streams consisting of different geometric representations of the same skeleton data,
while we are concerned with streams that correspond to different body parts.}  For example, prior work incudes multi-stream models for fingerspelling recognition (which combines hand and mouthing gestures)~\cite{shi2023toward}, SLT~\cite{ camgoz2020multi, zhou2021spatial, chen2022two, shi2022open, gueuwou2025signmusketeers},  and ISLR~\cite{pu2016sign, jiang2021skeleton}.  This factorization into multiple streams can enable dramatic improvements in data and compute efficiency over single-stream models that use the full image~\cite{gueuwou2025signmusketeers}.

Like this prior work, SHuBERT also adopts the idea of multiple streams.  However, unlike prior approaches, SHuBERT learns a \emph{self-supervised} representation from the multiple streams jointly that performs well on multiple tasks.

\begin{figure}[bt!]
    \centering
    \resizebox{\columnwidth}{!}{%
        \includegraphics{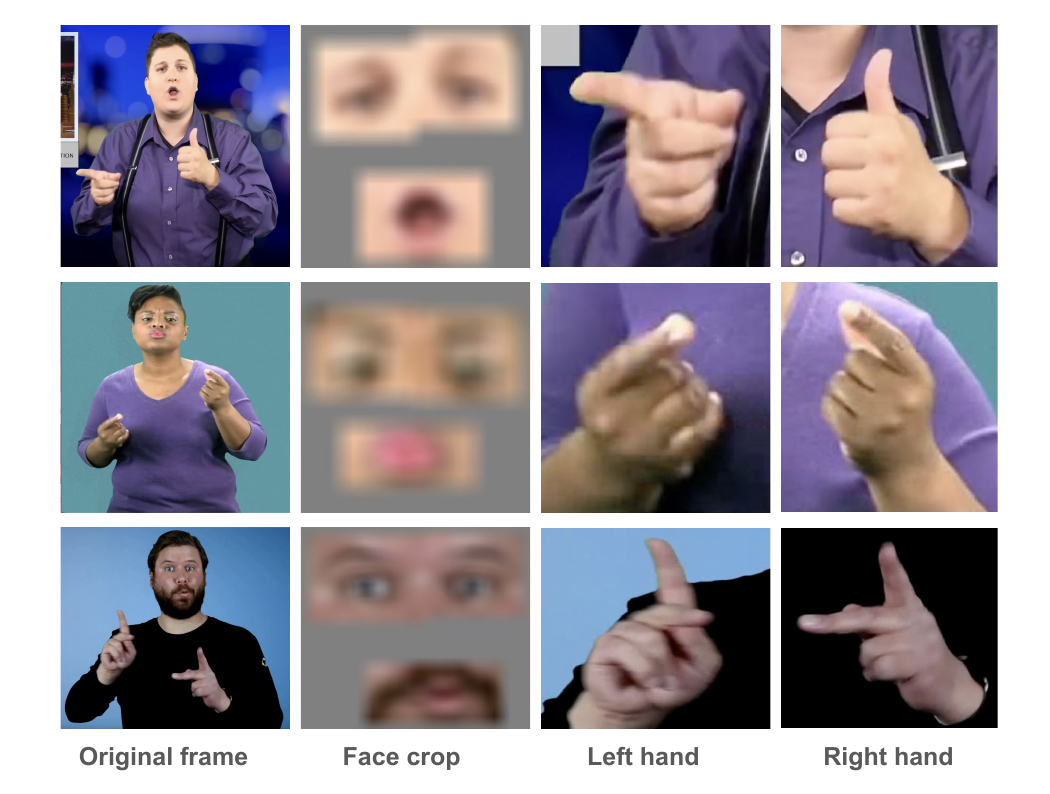}%
    }
    \vspace{-20pt}
    \caption{Sample frames of several signers and the corresponding input channels used for SHuBERT:  blurred face crop, left hand crop, and right hand crop. In addition to the face and left/right hand features extracted from these crops, each frame is represented by an additional feature vector corresponding to the upper body pose extracted from MediaPipe (see \cref{fig:shubert_pipeline}).}
    \label{fig:shubert_crops}
    \vspace{-10pt}
\end{figure}

\section{Sign Hidden-Unit BERT (SHuBERT)}
\label{sec:method}

SHuBERT is a transformer encoder~\cite{vaswani2017attention} that learns contextualized representations of sign language video frames through self-supervised learning.  The pre-training approach is outlined in \cref{fig:shubert_pipeline}.
In the following sections we describe the video features used in SHuBERT (\cref{sec:getting_the_components}) and the self-supervised training approach (\cref{sec:shubert}).

\subsection{Multi-Stream Feature Pre-Processing}
\label{sec:getting_the_components}

\cref{fig:shubert_crops} provides examples of SHuBERT's input video features, described in detail below.

\paragraph{Handshapes}
We use the MediaPipe Hand Landmarker\footnote{\url{https://ai.google.dev/edge/mediapipe/solutions/vision/hand_landmarker}} model,
which has hand detection accuracy $\sim$95\% on OpenASL.\footnote{Estimated from a small-scale experiment on 100 OpenASL videos, with manual verification of detections.} Upon inspection, we find that the majority of the remaining 5\% of ``failed" detections occur when the hands are outside the frame.
For these cases, we interpolate from the nearest frames with successful detections. Dilated bounding boxes for the detected hand landmarks (for both left and right hand) are cropped and resized to 224\texttimes224.

\paragraph{Facial Features} The signer's face contains important non-manual markers for sign languages~\cite{bragg2019sign}. 
Previous approaches either use the full face, compromising privacy \cite{gueuwou2025signmusketeers}, or %
blur the whole face in an attempt to protect privacy, potentially losing essential non-manual markers \cite{rust-etal-2024-towards}. Our design attempts to balance the need to preserve linguistic information 
with the goal of enhancing 
privacy. We identify the whole face, mouth and eye regions in the frame from the relevant MediaPipe facial landmarks. The face pixels are greyed out \emph{except for the pixels in the mouth and eye regions}. 
We then apply Gaussian blur to the entire face region and %
resize it to 224\texttimes224.

\paragraph{Image Feature Extractor for Hands and Face}
We use DINOv2~\cite{oquab2024dinov}, which has proven successful in previous  sign language work \cite{wong2024signgpt, gueuwou2025signmusketeers}, as the feature extractor for face and hand image crops.
An additional benefit of DINOv2 representations is that  
 they yield meaningful clusters after quantization~\cite{zheng2024stylebreeder}, which is an important property since SHuBERT training targets are clustered input features. While 
 many other prior approaches use keypoint estimation tools, 
 these have some weaknesses 
 in capturing 
 handshapes \cite{moryossef2021pose} and facial expressions \cite{kuznetsova2024mediapipe}. 

To adapt the general image feature extractor to a face feature extractor for sign language, we randomly sample 5 million face crops from videos in the YouTube-ASL \cite{uthus2023youtube} dataset and use them for continued pre-training of DINOv2 for 1 epoch. We do the same for the hand feature extractor, using 5 million randomly sampled hands (mix of left and right hand crops) from YouTube-ASL. %
For both the face and hand streams, crop regions of interest (ROIs) are processed through the face or hand fine-tuned DINOv2 models, yielding a 384-dimensional feature vector per crop, which we denote $\mathbf{x}^f_t, \mathbf{x}^l_t, \mathbf{x}^r_t\in\mathbb{R}^{384}$ for the face, left hand, and right hand features respectively.

\paragraph{Body Pose}
For (coarse) body pose, we extract seven key upper body landmarks (nose, shoulders, elbows, and wrists) and normalize their coordinates relative to the signing space, resulting in a compact 14-dimensional pose vector, $\mathbf{x}^b_t\in\mathbb{R}^{14}$.

\subsection{Self-Supervised Training of SHuBERT}
\label{sec:shubert}

\begin{figure}[tb!]
    \centering
    \includegraphics[width=\linewidth]{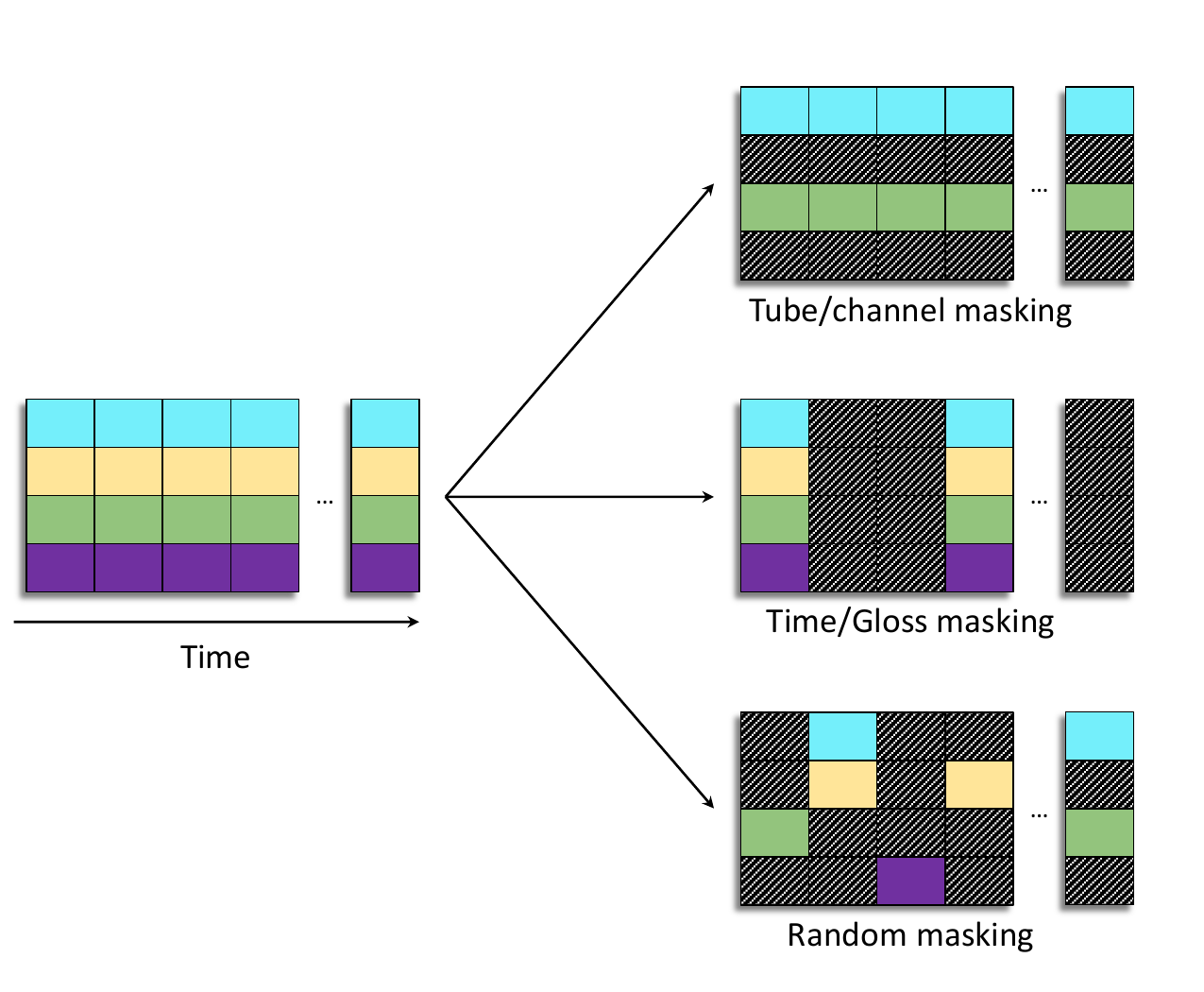}
    \vspace{-10pt}
    \caption{Three 
    strategies for sequence masking.  Each color-coded row corresponds to one of the four channels (face, right hand, left hand, body).}
    \label{fig:shubert_masking}
    \vspace{-10pt}
\end{figure}

For each of the four channels (left/right hand, face, and pose), we layer-normalize the extracted features  (producing zero mean and unit variance across all dimensions for each feature vector) and linearly project them to 256 dimensions, producing a 1024-dimensional input feature vector per frame.  This joint representation of the four streams is masked (see below) and input to the transformer encoder.
The output of the transformer 
is one representation vector per input frame, $y_t$, which may 
take into account information from the entire length of the input video, and from all four input channels.

\paragraph{Sequence Masking}
SHuBERT learns by predicting %
masked elements of the input feature sequence, given the observed (unmasked) data. 
We use a masking strategy designed for \emph{multi-channel} sign language input. We consider three types of masking, illustrated in \cref{fig:shubert_masking}: channel masking, which masks entire channels (e.g., all face and left hand features in a video) to learn cross-channel dependencies; time masking, which masks all channels at selected temporal positions (e.g., face, hand, and body pose in frames 20-40 in a given video); and random masking, which independently masks random small frame spans in each channel.  Based on our experiments comparing these strategies (\cref{sec:ablations}), we ultimately chose random masking.

\paragraph{Learning Objective}  
We use offline $k$-means clustering (separately for each of the four channels)
to create discrete target pseudo-labels $(f_t, l_t, r_t, b_t)$ for the face, left hand, right hand, and body pose respectively. The transformer output vector for each frame is fed to four linear classifiers (one per channel) to predict the cluster assignments for masked channels for that frame. 

As an example, suppose that face and body pose channel features for frame $t$, respectively $\mathbf{x}^f_t\in\mathbb{R}^{384}$ and $\mathbf{x}^b_t\in\mathbb{R}^{14}$, are masked. The $k$-means cluster assignments for these masked feature vectors are, respectively, $f_t$ and $b_t$, each a number between 1 and $k$. The classifier predicts, from the output vector $y_t$ for frame $t$, labels $\widehat{f}_t$ and $\widehat{b}_t$. 
The training objective is a cross-entropy loss between the target and predicted cluster assignments for the masked positions. The unmaksed positions are not included in the loss (but of course influence the predictions for the masked ones).

This self-supervised training produces the pre-trained SHuBERT model, which can then be fine-tuned for downstream sign language tasks using appropriate task-specific prediction layers and losses.

\section{Experiments and Results}
\label{sec:setup}
In this section, we describe the experimental setup for self-supervised training of SHuBERT, followed by its adaptation as a foundation model for multiple sign language processing tasks: sign language translation (\cref{sec:setup_downstream_translation}), isolated sign language recognition (\cref{sec:setup_downstream_recogntion}) and fingerspelling detection (\cref{sec:setup_downstream_fingerspelling_detection}). %
We also apply SHuBERT to phonological feature recognition, as a baseline for future work (see \cref{sec:setup_downstream_phono}).

\subsection{Pre-Training SHuBERT}
\label{sec:setup_pre-training}

\paragraph{Data and Pre-Processing}
For pre-training SHuBERT, we use the YouTube-ASL dataset \cite{uthus2023youtube}.
Note that {\it we excluded the clips that intersect with the OpenASL dataset}
\cite{shi2022open} to evaluate SHuBERT on OpenASL. 
To maintain the same training set size as the original YouTube-ASL dataset, we %
replaced the removed content with ASL videos from YouTube-SL-25~\cite{tanzer2024youtubesl25} that are not present in YouTube-ASL. Our final pre-training dataset comprises approximately 984 hours of ASL content.\footnote{Note that YouTube-ASL encompasses parts of several datasets used in other work, including 
72.4\% of the OpenASL test set and 
38.2\% of the MSASL~\cite{joze2018ms} test set.  We do not compare to other work on MSASL for this reason.}

To reduce computation, we downsample the videos by removing every other frame \cite{uthus2023youtube}. The average frame rate of the post-processed videos is 14.89 fps.

\paragraph{Model Configuration and Training}
We train a base model consisting of 12 transformer blocks, with each block having an embedding dimensionality of 768, feed-forward dimensionality of 3072, and 12 attention heads. The complete model contains 86M parameters. For each channel (face, left hand, right hand, and body pose), we use $k$-means on 10\% of the data to create 256 discrete clusters that serve as prediction targets.  \cref{fig:cluster_face,fig:cluster_lhand,fig:cluster_rhand,fig:cluster_body} in the Appendix provide examples of images in several clusters.
Our masking strategy uses a span length of 3 frames (approximately 200ms), 
which is roughly the average duration of fingerspelling a single letter in ASL and is therefore roughly the smallest gesture length
~\cite{hanson1982orthographic}.
We train the model for 400K steps using 8 NVIDIA A6000 (48GB) GPUs, with a total training time of approximately 7 days.

We optimize the model using Adam with a peak learning rate of $5 \times 10^{-4}$, warming up for the first $8\%$ of updates followed by linear decay. We batch videos to maintain efficiency while not exceeding 1,500 frames per GPU.

\subsection{Sign Language Translation}
\label{sec:setup_downstream_translation}
For sign language translation, where the input sign language video is mapped to text in English, we 
use ByT5-Base~\cite{xue2022byt5}, pre-trained on a large corpus of unlabeled multilingual text data, as a translation model to map from SHuBERT representations to written English, following prior work showing its strong performance on this task~\cite{tanzer2024youtubesl25, zhang2024scaling}.
We first extract video representations from SHuBERT and project them to ByT5's input space through a 
linear layer. 
We train the combined model (SHuBERT+projection layer+ByT5) with a cross-entropy loss and label smoothing factor of 0.2.

Similarly to \citet{uthus2023youtube} and \citet{rust-etal-2024-towards}, we use a two-phase training strategy. In the first phase, we train the translation system on the weakly labeled 
YouTube-ASL dataset (with OpenASL removed, as described above) for 250K steps. During this phase, we use the AdamW optimizer with a peak learning rate of $5\times 10^{-4}$ for ByT5 and a reduced learning rate of $5\times10^{-5}$ for SHuBERT parameters (when fine-tuned). The learning rate follows a cosine schedule with 10K warmup steps. We use a batch size of 2 utterances per GPU with gradient accumulation over 8 steps and use weight decay of 0.1.

In the second phase, we fine-tune on two target benchmark datasets (How2Sign~\cite{duarte2021how2sign} and OpenASL~\cite{shi2022open}) for 50K steps, using a lower learning rate of $10^{-4}$ and 5K warmup steps. We also evaluate in a zero-shot setting, without any additional fine-tuning, on a third dataset (for which no training data exists), FLEURS-ASL~\cite{tanzer2024fleurs}.
We use a learned weighted sum of features from all SHuBERT layers rather than using a single layer's output, as is commonly done when using speech representations such as HuBERT~\cite{yang2021superb}.
During decoding, we use beam search with a beam width of 5 and a maximum sequence length of 384 tokens.

\begin{table*}[t]
\begin{threeparttable}

\resizebox{1.0\textwidth}{!}{
\setlength{\tabcolsep}{5pt}
\begin{tabular}{@{}l r r | r r | r r | r r@{}}
\toprule
\multirow{2}{*}{Method} & \multirow{2}{*}{SSL} & \multirow{2}{*}{PT data (hrs)} & \multicolumn{2}{c|}{How2Sign} & \multicolumn{2}{c|}{OpenASL} & \multicolumn{2}{c}{FLEURS-ASL} \\
& & & BLEU$\uparrow$ & BLEURT$\uparrow$ & BLEU$\uparrow$ & BLEURT$\uparrow$ & BLEU$\uparrow$ & BLEURT$\uparrow$ \\
\midrule
\addlinespace[0.5em]
\rowcolor{gray!20}\multicolumn{9}{c}{\bf Private data} \\
\addlinespace[0.5em]
~~\citet{tanzer2024fingerspelling} & $\times$ & $\sim$2,800 & 18.1 & 50.8 & - & - & {\bf 5.8} & {\bf 45.4} \\
~~\citet{zhang2024scaling} & $\times$ & $\sim$6,600 & {\bf 21.1} & {\bf 55.7} & - & - & - & - \\
\addlinespace[0.5em]
\rowcolor{gray!20}\multicolumn{9}{c}{\bf Publicly available data} \\
\addlinespace[0.5em]
~~SSVP \cite{rust-etal-2024-towards} & $\checkmark$ & 1,054 & 15.5 & 49.6 & - & - & - & - \\
~~\citet{uthus2023youtube} & $\times$ & 984 & 12.4 & 46.6 & - & - & - & - \\
~~\citet{tanzer2024youtubesl25} & $\times$ & 3,207 & 15.4 & 47.9 & - & - & 4.4 & 40.1 \\
~~SM \cite{gueuwou2025signmusketeers} & $\checkmark$ & 984 & 14.3 & - & - & - & - & - \\
~~VAP \cite{jiao2024visual} & $\times$ & - & 12.9 & - & 21.2 & - & - & - \\
~~Uni-Sign \cite{li2025unisign} & $\times$ & 984 & 14.9 & 49.4 & 23.1\tnote{*} & 60.4\tnote{*} & - & - \\
~~OpenASL \cite{shi2022open} & $\times$ & - & - & - & 6.7 & 31.1 & - & - \\
~~GloFE-VN \cite{lin2023gloss} & $\times$ & - & - & - & 7.1 & 36.7 & - & - \\
~~C2RL \cite{chen2025c2rl} & $\times$ & - & - & - & 13.2 & - & - & - \\
~~\citet{tanzer2024fleurs} & $\times$ & 984 & - & - & - & - & 3.9 & 38.3 \\
\midrule
~~Ours & $\checkmark$ & 984 & {\bf 16.2} & {\bf 49.9} & {\bf 23.2} & {\bf 60.6} & {\bf 4.7} & {\bf 41.2} \\
\bottomrule
\end{tabular}
}
\caption{\label{tab:translation_results_combined}Translation results
on the How2Sign, OpenASL, and FLEURS-ASL test sets. SSL: self-supervised learning (yes/no); PT: pre-training.
* Uni-Sign is pre-trained on YouTube-ASL, which contains >72\% of the OpenASL test samples, so we do not consider the results on OpenASL to be directly comparable to ours.}
\end{threeparttable}

\end{table*}

We evaluate the final model on the three benchmark test sets of How2Sign, OpenASL, and FLEURS-ASL, using 
the standard BLEU \cite{papineni2002bleu,post2018call}\footnote{\url{SacreBLEU version signature: BLEU+c.mixed+\#.1+s.exp+tok.13a+v.1.4.1.}} and BLEURT \cite{sellam2020bleurt} translation metrics, as shown in \cref{tab:translation_results_combined} (see also example translations in \cref{tab:supp+h2s,tab:supp+fleursasl,tab:supp+oasl}).
 
 In all cases, our results using SHuBERT improve over the best prior published results using publicly available training data.  On How2Sign, 
 SHuBERT improves by +0.7 BLEU/+0.3 BLEURT over the prior state-of-the-art result (using public data) of SSV-SLT~\cite{rust-etal-2024-towards}, which used slighty more pre-training data (1,054 vs.~984 hours) that
 included the training data of How2Sign.
 Better published results exist (as shown in \cref{tab:translation_results_combined}), but they rely on private fine-tuning datasets so we cannot reproduce their settings nor compare to them meaningfully.  

In the case of OpenASL, SHuBERT's improvement over the best prior result is +2.0 BLEU.\footnote{Uni-Sign \cite{li2025unisign} reports a similar BLEU score of 23.1 on OpenASL, but Uni-Sign is pre-trained on YouTube-ASL.  Therefore, most of the test set is included in Uni-Sign's pre-training data, so we do not consider the results directly comparable.}
This larger improvement may be attributable to pre-training on similar-domain (generally native, natural rather than interpreted signing) data in YouTube-ASL (but, as previously mentioned, none of the same data). The distinction between interpreted signing in a constrained visual environment (as in How2Sign and FLEURS-ASL) and natural signing in a less constrained environment (as in OpenASL and much of YouTube-ASL) is an important one that has not received sufficient attention,
and is worth exploring further in future work.  The properties of the visual environment affect the difficulty of the task, and natural signing has different characteristics from those of interpreted sign language~\cite{desai2024systemic}.

On FLEURS-ASL, a dataset designed for testing only, SHuBERT demonstrates strong performance in a \emph{zero-shot} setting, surpassing both prior methods, which used over 3x as much pre-training data (3,207 hours) as ours. 

We also note that some previous approaches use additional 
techniques
such as auxiliary losses
that contribute to their final results, such as additional contrastive learning with labelled data \cite{rust-etal-2024-towards}, joint training with text machine translation \cite{zhang2024scaling},  and multi-tasking with random dynamic clips from an original video \cite{tanzer2024fleurs}. 
It is possible that incorporating such techniques into our framework will further improve performance, but we leave this for future work. 

\vspace{.1in}
\noindent\textbf{Ablations} We conduct
several ablation studies to validate our design choices and analyze SHuBERT's behavior. These studies examine: (1) the impact of different masking strategies during pre-training, where random masking proves most effective based on BLEURT scores; (2) the importance of pre-training data scale, showing the clear benefit of using the full pre-training dataset;
(3) the contribution of different architectural components, demonstrating that a weighted combination of layers significantly improves translation performance; and (4) the effects of fine-tuning versus keeping SHuBERT frozen during translation training, with fine-tuning providing moderate gains. 
The strong performance of the frozen, layer-weighted SHuBERT suggests that it is a promising approach
for low-resource settings where parallel data may be limited. Detailed results and analysis of these ablations can be found in \cref{sec:ablations}.

\subsection{Isolated Sign Language Recognition }
\label{sec:setup_downstream_recogntion}

\begin{table*}
\centering

\vspace{-10pt}
\resizebox{\linewidth}{!}{
    \begin{tabular}{l  r | r r r | r r r | r r}
    \toprule
    \multirow{2}{*}{Method} & \multirow{2}{*}{\#Params} & \multicolumn{3}{c|}{ASL Citizen} & \multicolumn{3}{c|}{Sem-Lex} & \multicolumn{2}{c}{WLASL2000} \\
    & & Rec@1$\uparrow$ & Rec@5$\uparrow$ & Rec@10$\uparrow$ & Rec@1$\uparrow$ & Rec@5$\uparrow$ & Rec@10$\uparrow$ & P-I$\uparrow$ & P-C$\uparrow$ \\
    \midrule
    ST-GCN \cite{desai2024asl} & 0.45M & 0.60 & 0.82 & 0.88 & - & - & - & - & - \\
    SignCLIP \cite{jiang2024signclip} & 217M & 0.60 & 0.84 & 0.89 & 0.30 & 0.48 & 0.55 & - & - \\
    I3D \cite{desai2024asl} & 25M & 0.63 & 0.86 & {\bf 0.91} & - & - & - & - & - \\
    Sem-Lex \cite{kezar2023sem} & 0.45M & - & - & - & $0.69^*$ & - & - & - & - \\
    SignBERT \cite{hu2021signbert} & - & - & - & - & - & - & - & 39.40 & 36.74 \\
    SignBERT+ \cite{hu2023signbert+} & - & - & - & - & - & - & - & 48.85 & 46.37 \\
    MSLU \cite{zhou2024multimodal} & - & - & - & - & - & - & - & 56.29 & 53.29 \\
    NLA-SLR \cite{zuo2023naturalsign} & - & - & - & - & - & - & - & 61.05 & 58.05 \\
    Uni-Sign \cite{li2025unisign} & 580M & - & - & - & - & - & - & {\bf 63.52} & {\bf 61.32} \\
    \midrule
    Ours (rank=1 LoRA) & 0.17M & {\bf 0.65} & {\bf 0.87} & {\bf 0.91} & {\bf 0.54} & {\bf 0.74} & {\bf 0.80} & 60.90 & 58.01 \\
    \bottomrule
    \end{tabular}
}
\small
\caption{ISLR %
results on the ASL Citizen, Sem-Lex, and WLASL2000 test sets. Note: For Sem-Lex, the result marked with an asterisk ($^*$) is not directly comparable to Ours as it is for a reduced (and easier) test set, as mentioned in \cite{jiang2024signclip}. Additionally, the dataset version released by \cite{kezar2023sem} has a significant fraction of videos missing. For WLASL2000, evaluation metrics are per-instance (P-I) and per-class (P-C) Top-1 accuracy.}
\label{tab:islr_results_combined}
\end{table*}

For isolated sign language recognition (ISLR), the task of classifying a short video of a single sign,
we 
include results for SHuBERT adapted with LoRA adapters~\cite{hu2022lora}. 
Unless otherwise specified, for all experiments, we train for 125 epochs with a batch size of 128 and perform early stopping according to  validation results (R@1/P-I). We use an Adam optimizer \cite{kingma2014adam} with a learning rate of $10^{-4}$ and weight decay of $10^{-4}$. For %
classification tasks, we first average SHuBERT representations across the time dimension and add a batch-norm layer followed by a linear layer as the classification head. 

For LoRA training, we learn a rank-1 LoRA module for each linear layer in SHuBERT while keeping all the other weights frozen, resulting in training only %
0.2\% of the number of parameters of the original model. In addition to the aforementioned hyperparameters, we reduce the learning rate of the LoRA modules to 1/10 of the classification head's and also use 0.1 label smoothing.

Our ISLR results on ASL Citizen,  WLASL2000 (original), and Sem-Lex are shown in \cref{tab:islr_results_combined}.
We note that we do not report on MSASL, as done in some of the prior work, because of the aforementioned overlap between %
its test set 
and YouTube-ASL.  Following prior work, we report Recall at 1, 5 and 10, unless stated otherwise.\footnote{Note that some prior work reports ISLR results in terms of accuracy, which is equivalent to Recall at 1.} We achieve state-of-the-art performance on all datasets except WLASL2000 where Uni-Sign has a better result.  We note that Uni-Sign fine-tunes 3,000 times more paramters than ours.\footnote{Note:  As described in the caption of \cref{tab:islr_results_combined}, the Sem-Lex results in~\cite{kezar2023sem} are not comparable with other methods, including ours.}

\subsection{Fingerspelling detection}
\label{sec:setup_downstream_fingerspelling_detection}
\begin{table}
\small
    \centering

    \vspace{-10pt}
    \setlength{\tabcolsep}{2.5pt} 
    \begin{tabular}{l r r }
        \toprule
        Method & SSL & Mean IoU $\uparrow$ \\
        \midrule
        Contrastive Learning \cite{yin2024asl} & $\times$ & 0.28 \\
        SHuBERT (Ours) & $\checkmark$ & {\bf 0.40} \\
        \bottomrule
    \end{tabular}
\small
    \caption{%
    Fingerspelling detection on ASL-Stem Wiki \cite{yin2024asl}.}
\label{tab:fingerspelling_aslstemwiki}
\end{table}

Finally, we evaluate SHuBERT on the task of %
fingerspelling detection on the ASL-Stem-Wiki dataset~\cite{yin2024asl}. Given a sign language video input $v$, which consists of an ordered sequence of frames $\{v_1, v_2, \ldots, v_n\}$, 
the task is to identify all segments containing fingerspelling. The output is represented as a set $F$ of frame intervals: $F = \{[s_1, e_1], [s_2, e_2], \ldots, [s_k, e_k]\}$, where each interval $[s_i, e_i]$ represents the start and end frames of a fingerspelling sequence, such that frames $v_{s_i}$ through $v_{e_i}$ contain fingerspelling. 
We follow the original 
ASL-Stem-Wiki evaluation pipeline (cross-validation) and evaluation metric (intersection over union, or IoU). The results 
are shown in \cref{tab:fingerspelling_aslstemwiki}. We see that by simply fine-tuning SHuBERT for this task, we increase the IoU by 42\% compared to the previous state of the art method, which pre-trains and fine-tunes on the same dataset.

\section{Conclusion}
\label{sec:conclusion}

SHuBERT, our proposed self-supervised approach for learning sign language video representations,
yields a transformer encoder that maps from multiple feature streams (face and hand appearance and upper body pose) to a stream of per-frame contextual representations.  A single base SHuBERT model, when adapted to a range of sign language processing benchmarks including both translation and isolated sign recognition, achieves strong performance on all of them and almost always improves over the prior state of the art. 
SHuBERT is trained on public data and is publicly available.\footnote{\url{http://shubert.pals.ttic.edu}} 
 Based on its strong performance on the tasks studied here, we expect that SHuBERT can serve as a base model for a broad range of sign language processing tasks.  

\newpage
\paragraph{Limitations}
Our work has several limitations.  First, although the results are competitive with or outperform prior work, the absolute performance is still quite poor.  Neither our model nor others can replace human interpreters for broad-domain sign language translation. Second, our training data volume is significantly smaller than that of 
typical self-supervised speech and text models. We cannot say with certainty how our model would scale up to much larger datasets.  Third, we have not carefully studied potential sources of bias in the model. 
 From our qualitative visual inspection of images and their corresponding clusters (\cref{fig:cluster_face,fig:cluster_lhand,fig:cluster_rhand,fig:cluster_body} in the Appendix), we observe that semantic properties appear to take precedence over attributes like skin color, gender, or eyewear. While these preliminary observations are encouraging, a more thorough investigation of potential biases would be valuable future work. 
 Finally, the current scope of our work is limited to American Sign Language. Although sign languages use the same channels and share many elements, we do not know how well our model would generalize to other languages.  Future work could address these limitations by expanding the training dataset and training on data from 
other sign languages.

\paragraph{Acknowledgment}
We are grateful to Shiry Ginosar, Anand Bhattad, Ju-Chieh Chou, and Chung-Ming Chien for their valuable suggestions throughout this project.

\bibliography{custom}

\appendix

\clearpage
\setcounter{page}{1}

\section{Ablations on SHuBERT}\label{sec:ablations}

We conduct ablations, using the How2Sign translation task, to measure the importance of different factors in the SHuBERT pre-training and adaptation to the downstream task. 
Unless stated otherwise, in the ablation experiments we pre-train SHuBERT for 100K steps (instead of the full 400K) for a faster turnaround and freeze SHuBERT for the downstream task.

\paragraph{Masking Strategies}
\begin{table}[t]
    \centering

    \vspace{-10pt}
    \resizebox{1.0\linewidth}{!}{%
    \begin{tabular}{l r r r}
        \toprule
        Masking strategy & BLEU-1 & BLEU & BLEURT \\
        \midrule
        Channel masking & 14.5 & 2.6  & 29.9\\
        Time masking & 15.1 & 2.3 & 31.2\\
        Random masking  & 15.4 & 2.2 & 31.4\\
        \bottomrule
    \end{tabular}
        }
            \caption{%
    Comparison of masking strategies in SHuBERT pre-training. 
 SHuBERT is frozen and stopped after 100K steps, and fine-tuned and evaluated on How2Sign only.}
\label{tab:masking_strategy_shubert}
\end{table}

  \cref{tab:masking_strategy_shubert} compares translation performance when using each of the three masking strategies (\cref{fig:shubert_masking}). 
 We observe different signals from different metrics. While the BLEU scores suggest channel masking to be the best, random masking produces better BLEURT scores. Contradicting signals between different evaluation metrics for sign language translation has also been observed in prior work \cite{zhang2024scaling}. We chose to priotize BLEURT, as it generally has better alignment with human judgements of translation quality~\cite{freitag2022wmt}, and therefore use random masking in all of our other experiments.

\paragraph{Data Scaling Behavior of SHuBERT}
\begin{table}[t]
   \centering
   \vspace{-10pt}
   \resizebox{1.0\linewidth}{!}{%
   \begin{tabular}{r r r r}
       \toprule
       Hours of pre-train data & BLEU-1 & BLEU & BLEURT \\
       \midrule
       984 & 15.4 & 2.2 & 31.4 \\
       98 & 12.7 & 0.7 & 29.1 \\
       \bottomrule
   \end{tabular}
   }
      \caption{Impact of pre-training data size on SHuBERT's performance.}
\label{tab:scaling_shubert}
\end{table}

In addition to the full pre-training dataset, we also train SHuBERT from scratch on a randomly selected 10\% of the full data. In \cref{tab:scaling_shubert} we see that there is a noticeable drop in performance in BLEU and BLEURT, suggesting that data size is important. With multiple larger  datasets now available---BOBSL~\cite{Albanie2021bobsl}, JWSign~\cite{gueuwou2023jwsign} and YouTube-SL-25~\cite{tanzer2024youtubesl25} contain approximately 1500, 2500, and 3200 hours of data respectively---we expect that expanding the SHuBERT pre-training data may further improve performance.  In addition, these larger datasets also include more language diversity, which may also improve performance and/or applicability to additional languages.

\vspace{-.1in}
\paragraph{Isolating SHuBERT's Impact on Performance}
\begin{table}[t]
   \centering

   \vspace{-10pt}
   \resizebox{1.0\linewidth}{!}{%
   \begin{tabular}{l r r r}
       \toprule
       Layer of SHuBERT & BLEU-1 & BLEU & BLEURT \\
       \midrule
       None & 15.3 & 2.5 & 31.6 \\
       Last Layer & 21.4 & 4.7 & 35.0 \\
       Weighted Sum & 29.3 & 7.1 & 39.5 \\
       \bottomrule
   \end{tabular}
   }
      \caption{Contribution of several components of SHuBERT:
   direct use of input video features (None) vs.~using SHuBERT's last layer vs.~weighted combination of all layers. In all cases SHuBERT is frozen.}
\label{tab:shubert_necessary}
\end{table}

To understand SHuBERT's impact on translation performance, we conduct three experiments, shown in \cref{tab:shubert_necessary}. In our baseline experiment (``None"), we directly feed the projected 4-channel features (face, left hand, right hand, body pose) to the ByT5 translation model, bypassing SHuBERT entirely, resulting in fairly poor performance. When we instead pass these features through a frozen pre-trained SHuBERT (400k steps, random masking) and use its final layer's output (``Last Layer"), we see significant improvement. Finally, computing a learned weighted combination of all SHuBERT layers (``Weighted Sum") further improves performance. These results demonstrate that each component of SHuBERT 
contribute to translation performance.

\paragraph{Frozen vs.~Fine-Tuned SHuBERT}
\begin{table}[t]
   \centering

   \vspace{-10pt}
   \resizebox{1.0\linewidth}{!}{%
   \begin{tabular}{r r r r}
       \toprule
       Fine-tune SHuBERT & BLEU-1 & BLEU & BLEURT \\
       \midrule
       $\times$ & 21.4 & 4.7 & 35.0 \\
       $\checkmark$ & 30.0 & 7.5 & 39.9 \\
       \bottomrule
   \end{tabular}
   }
      \caption{Effect of fine-tuning on translation performance: Fine-tuning SHuBERT along with the translation model ($\checkmark$) vs.~using frozen SHuBERT representations ($\times$).}
\label{tab:finetune_shubert}
\end{table}

We also investigate the impact of fine-tuning during translation training, as shown in \cref{tab:finetune_shubert}. We compare two scenarios: using a frozen SHuBERT (400k steps, random masking) and only fine-tuning ByT5, versus fine-tuning both SHuBERT (from the same base model) and ByT5 together. Both scenarios use features from SHuBERT's last layer. Fine-tuning SHuBERT leads to substantial improvements compared to keeping it frozen, when using the final layer. However, referring back to~\cref{tab:shubert_necessary}, the relatively {\it small} difference between the fine-tuned performance and that of the frozen and layer-weighted SHuBERT is noteworthy. 
 This observation is 
promising for low-resource sign languages, where we may have plentiful unlabeled video data for pre-training, but very limited parallel data for translation training.

\paragraph{Stream contributions in ASL-to-English translation}
\begin{table}
    \centering

    \vspace{-10pt}
    \setlength{\tabcolsep}{2.5pt} 
    \begin{tabular}{l r }
        \toprule
        Streams Used & BLEU $\uparrow$ \\
        \midrule
        Face only & 0.6 \\
        Hands only & 0.2 \\
        Upper body only & 0.8 \\
        Hands + Upper Body & 2.1 \\      
        All streams &  {\bf 2.4} \\
        \bottomrule
    \end{tabular}
\small
    \caption{%
    Stream contributions in ASL-to-English translation.}
\label{tab:streams_contribution}
\end{table}

To quantify the contribution of each input stream, we conduct a translation experiment with the How2Sign dataset (without pre-training), feeding the concatenation of the multiple streams directly to a language model (T5). The resulting BLEU scores are shown in~\cref{tab:streams_contribution}, showing that all of the streams contribute to translation.  These results should be interpreted as assessing the true relative importance of each stream, however, since this is a very small-scale experiment.

\section{Phonological Feature Recognition}
\label{sec:setup_downstream_phono}

\begin{table}[t]
    \centering

    \vspace{-10pt}
    \resizebox{1.0\linewidth}{!}
    {%
    \begin{tabular}{l r r}
        \toprule
        \multirow{2}{*}{Phonological Feature} & \multicolumn{2}{c}{Rec@1$\uparrow$}  \\ 
        & Sem-Lex & ASL Citizen \\
        \midrule
        Major Location                          & 0.8477 & 0.9022\\
        \rowcolor{Gray} Minor Location          & 0.7130 & 0.8000\\
        Second Minor Location                   & 0.7328 & 0.8118\\
        \rowcolor{Gray} Contact                 & 0.8684 & 0.9157\\
        Thumb Contact                           & 0.8474 & 0.8752\\
        \rowcolor{Gray} Sign Type               & 0.8464 & 0.9154\\
        Repeated Movement                       & 0.8265 & 0.8993\\
        \rowcolor{Gray} Path Movement           & 0.7275 & 0.7942\\
        Wrist Twist                             & 0.9058 & 0.9300\\
        \rowcolor{Gray} Selected Fingers        & 0.7953 & 0.8344\\
        Thumb Position                          & 0.8604 & 0.8819\\
        \rowcolor{Gray} Flexion                 & 0.7264 & 0.7773\\
        Spread                                  & 0.7942 & 0.8480\\
        \rowcolor{Gray} Spread Change           & 0.8160 & 0.8658\\
        Nondominant Handshape                   & 0.7632 & 0.8432\\
        \rowcolor{Gray} Handshape               & 0.6293 & 0.7080\\
        \midrule
        Average                                 & 0.7938 & 0.8502\\
        \bottomrule
    \end{tabular}
    }

    \caption{%
        Phonological feature recognition accuracy on two datasets, Sem-Lex and ASL Citizen. 
    }  
\label{tab:phono}
\vspace{-10pt}

\end{table}

We also conduct experiments on phonological feature recognition, that is the classification of linguistic features of signs, for two of the ISLR datasets. We report the Recall at 1 (prediction accuracy) for 16 commonly used phonological features (from ASL-LEX 2.0~\cite{sehyr2021asllex}) in \cref{tab:phono}. The training setups and hyperparameters are identical to those of the full fine-tuning method in \cref{sec:setup_downstream_recogntion}, except that we now train 16 classification heads simultaneously. We also find that removing weight decay gives a slight performance boost. 

To the best of our knowledge, no previous work has reported phonological feature recognition accuracies on the ASL Citizen dataset. Similarly, though the Sem-Lex authors \cite{kezar2023sem} report phonological feature prediction accuracies, they are not comparable to ours, which are computed on the entire test set available to the public. Thus, we hope that our results in \cref{tab:phono} can serve as a benchmark for future work.

\section{ASL Phonological Feature Classification Details}

American Sign Language (ASL) can be described through a set of phonological features, similarly to the description of spoken languages via features. These features capture the essential components of sign formation, including hand configuration, movement patterns, and spatial relationships. 
 \cref{tab:phono_classes} presents the number of classes for each of the phonological features we use in our phonological feature prediction analysis for Sem-lex and ASL-Citizen, and below we list the values of each feature.  This commonly used feature set is from ASL-LEX 2.0~\cite{sehyr2021asllex}).
 
\begin{table}[t]
    \centering

    \vspace{-10pt}
    \resizebox{1.0\linewidth}{!}
    {%
    \begin{tabular}{l r r}
        \toprule
        \multirow{2}{*}{Phonological Feature} & \multicolumn{2}{c}{Number of Classes}  \\ 
        & Sem-Lex & ASL Citizen \\
        \midrule
        Major Location                          & 5 & 6 \\
        \rowcolor{Gray} Minor Location          & 37 & 37 \\
        Second Minor Location                   & 37 & 38 \\
        \rowcolor{Gray} Contact                 & 2 & 2 \\
        Thumb Contact                           & 3 & 3 \\
        \rowcolor{Gray} Sign Type               & 6 & 6 \\
        Repeated Movement                       & 2 & 2 \\
        \rowcolor{Gray} Path Movement           & 8 & 8 \\
        Wrist Twist                             & 2 & 2 \\
        \rowcolor{Gray} Selected Fingers        & 12 & 12 \\
        Thumb Position                          & 2 & 2 \\
        \rowcolor{Gray} Flexion                 & 8 & 8 \\
        Spread                                  & 3 & 3 \\
        \rowcolor{Gray} Spread Change           & 3 & 3\\
        Nondominant Handshape                   & 56 & 57 \\
        \rowcolor{Gray} Handshape               & 58 & 58 \\

        \bottomrule
    \end{tabular}
    }
        \caption{Number of classes for each phonological feature represented in two ASL datasets, Sem-Lex and ASL Citizen.  Most features have the same number of classes across datasets, while a few features have values that don't appear in one of the datasets (for example, Second Minor Location has 37 classes that appear in Sem-Lex and 38 classes in ASL Citizen).}
\label{tab:phono_classes}
\vspace{-10pt}
\end{table}

\paragraph{Handshape} v, 5, y, h, open\_b, c, baby\_o, flat\_h, o, l, 1, a, open\_8, w, curved\_5, d, flatspread\_5, i, f, s, p, flat\_b, curved\_4, flat\_o, g, open\_e, 4, closed\_b, bent\_1, 3, flat\_horns, goody\_goody, flat\_m, bent\_v, flat\_1, r, 8, curved\_v, open\_h, curved\_1, horns, flat\_ily, flat\_n, bent\_l, stacked\_5, ily, e, flat\_v, curved\_l, spread\_open\_e, curved\_h, 7, closed\_e, t, flat\_4, open\_f, k, and spread\_e.

\paragraph{Nondominant Handshape} v, 5, y, none, open\_b, Dominance Condition Violation, B, 1, a, open\_8, C, s, h, o, flat\_b, curved\_5, p, c, S, closed\_b, 4, flat\_m, bent\_v, flat\_1, flat\_h, baby\_o, curved\_v, i, f, bent\_1, Symmetry Violation, flatspread\_5, flat\_o, curved\_1, open\_h, stacked\_5, g, l, bent\_l, 3, 8, spread\_open\_e, e, horns, w, r, Lax, curved\_l, open\_e, flat\_4, O, curved\_b, A, ily, flat\_v, and flat\_horns.

\paragraph{Minor Location} Neutral, Head Away, Body Away, Hand Away, Palm, Finger Tip, Forehead, Finger Front, Mouth, Chin, Other, Upper Arm, Torso Top, Forearm Back, Cheek Nose, Wrist Front, Palm Back, Finger Back, Finger Radial, Under Chin, Finger Ulnar, Wrist Back, Shoulder, Arm Away, Forearm Ulnar, Torso Mid, Heel, Clavicle, Eye, Forearm Front, Neck, Torso Bottom, Upper Lip, Head Top, Elbow Back, Hips, and Waist.

\paragraph{Second Minor Location} Neutral, Head Away, Torso Bottom, Finger Tip, Hand Away, none, Palm, Forearm Back, Finger Back, Body Away, Torso Top, Finger Front, Chin, Arm Away, Upper Arm, Finger Ulnar, Eye, Hips, Neck, Palm Back, Forearm Front, Finger Radial, Mouth, Heel, Torso Mid, Other, Waist, Cheek Nose, Forehead, Elbow Back, Under Chin, Clavicle, Shoulder, Forearm Ulnar, Head Top, Upper Lip, and Forearm Radial.

\paragraph{Sign Type} Symmetrical Or Alternating, One Handed, Dominance Violation, Asymmetrical Different Handshape, Asymmetrical Same Handshape, and Symmetry Violation.

\paragraph{Path Movement} Curved, Back And Forth, Straight, Circular, None, Z-shaped, Other, and X-shaped.

\paragraph{Flexion} Fully Open, Curved, Bent, Flat, none, Fully Closed, Stacked, and Crossed.

\paragraph{Selected Fingers} im, imrp, p, i, t, m, ip, imp, mr, imr, r, and mrp.

\paragraph{Major Location} Neutral, Head, Body, Hand, and Arm.

\paragraph{Flexion Change} 1.0, 0.0, and none.

\paragraph{Spread Change} 1.0, 0.0, and none.

\paragraph{Thumb Contact} 1.0, 0.0, and none.

\paragraph{Spread} 1.0, 0.0, and none.

\paragraph{Thumb Position} Closed and Open.

\paragraph{Repeated Movement} 1.0 and 0.0.

\paragraph{Contact} 1.0 and 0.0.

\paragraph{Wrist Twist} 0.0 and 1.0.

\section{Cluster Samples}
\label{sec:cluster_samples}
\begin{figure*}[t]
    \centering
    \includegraphics[width=\linewidth]{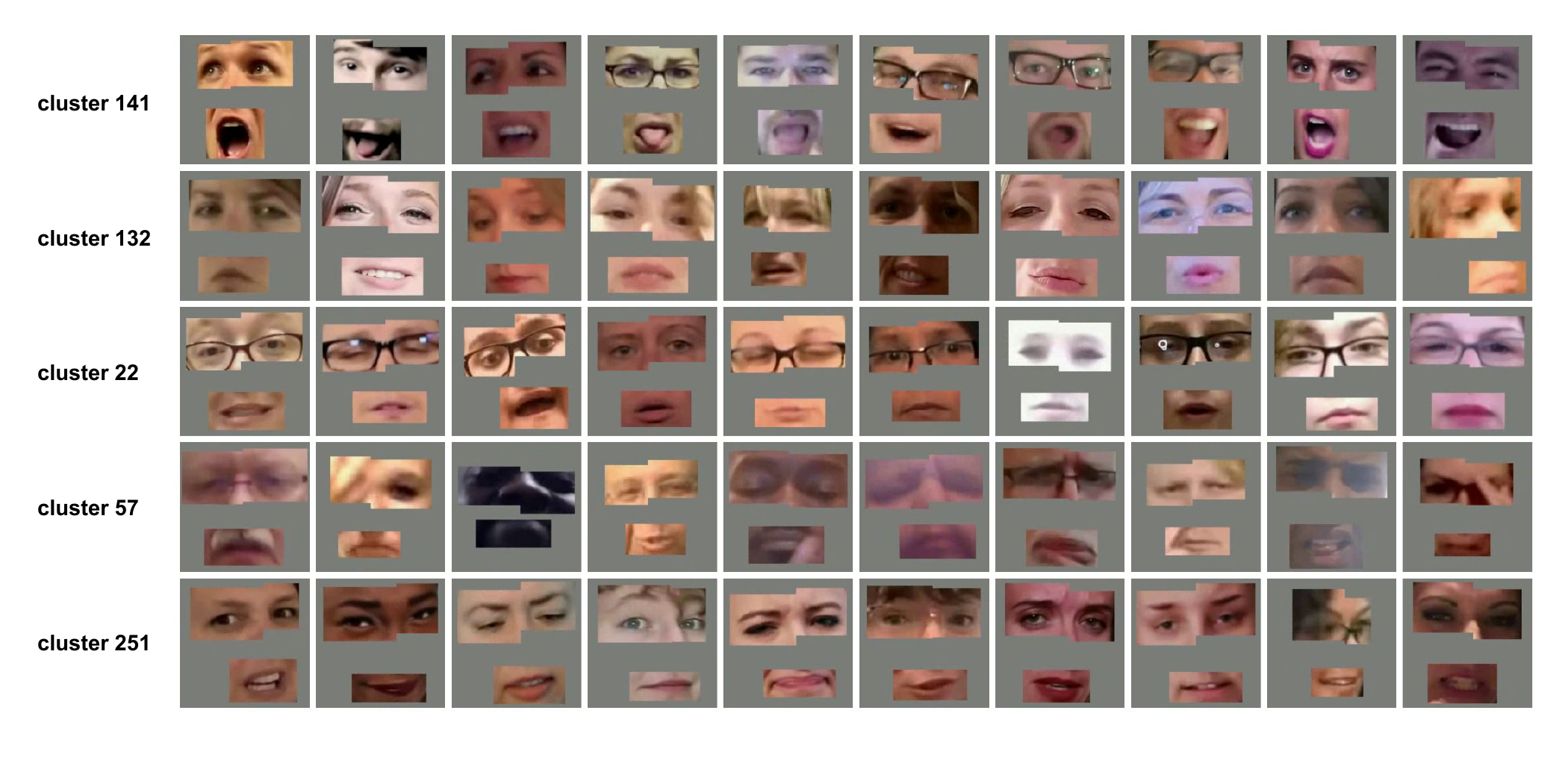}
    \caption{Sample face clusters.
    Each row represents a distinct cluster and 10 random examples from it. 
    Cluster 141 includes mainly open-mouthed expressions with raised eyebrows, 
    cluster 57 seems to capture closed or squinting eyes with neutral mouths, and cluster 251 corresponds to a slightly tilted head with direct gaze and little mouth opening. NOTE: For clarity, we show unblurred cropped faces here.}
    \label{fig:cluster_face}
\end{figure*}

\begin{figure*}[t]
    \centering
    \includegraphics[width=\linewidth]{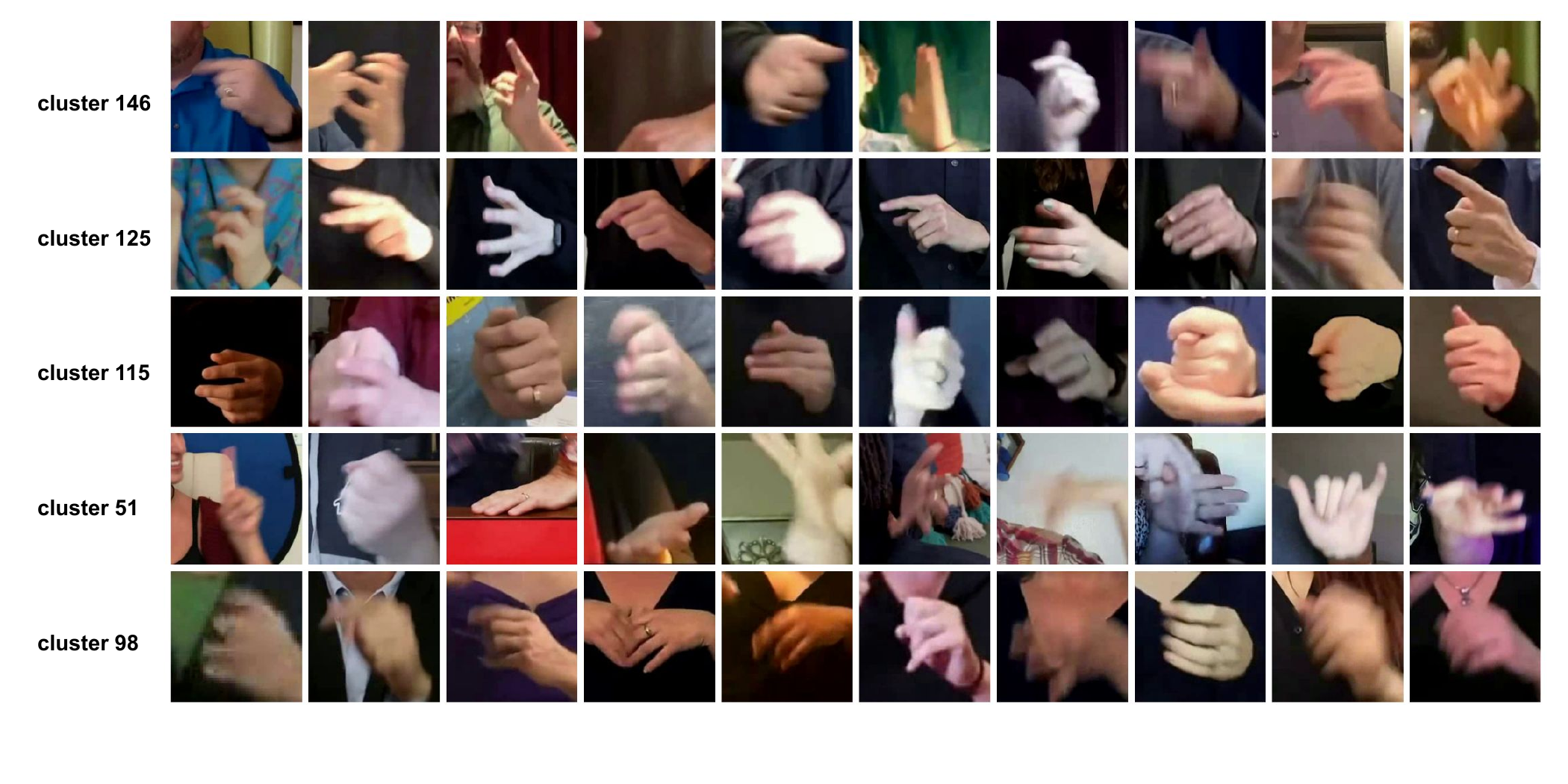}
    \caption{Sample left hand clusters.
    Each row represents a distinct cluster and 10 random examples from it. Cluster 125 shows pointing configurations with the index finger extended. Cluster 115 generally corresponds to closed fist formations oriented with the thumb on top. Cluster 51 seems to be a mix of poses without a consistent description.
    Cluster 98 seems to include mainly transitional hand movements around the chest.}

    \label{fig:cluster_lhand}
\end{figure*}

\begin{figure*}[t]
    \centering
    \includegraphics[width=\linewidth]{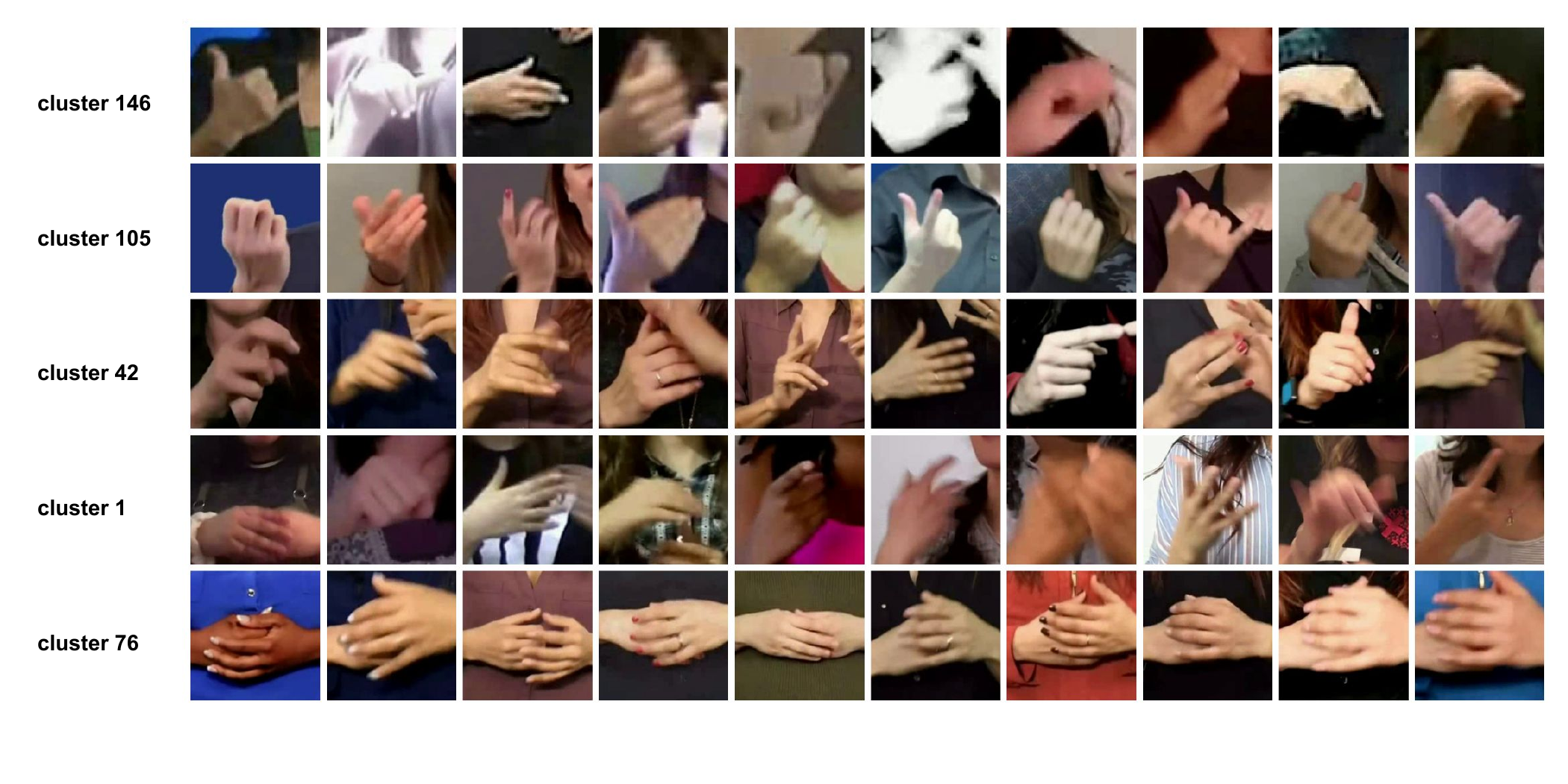}

    \caption{Sample right hand clusters.
    Each row represents a distinct cluster and 10 random examples from it. Cluster 42 captures mainly multi-finger pointing gestures.
    Cluster 76 corresponds to clasped or overlapped hands in resting positions.}

    \label{fig:cluster_rhand}
\end{figure*}

\begin{figure*}[t]
    \centering
    \includegraphics[width=\linewidth]{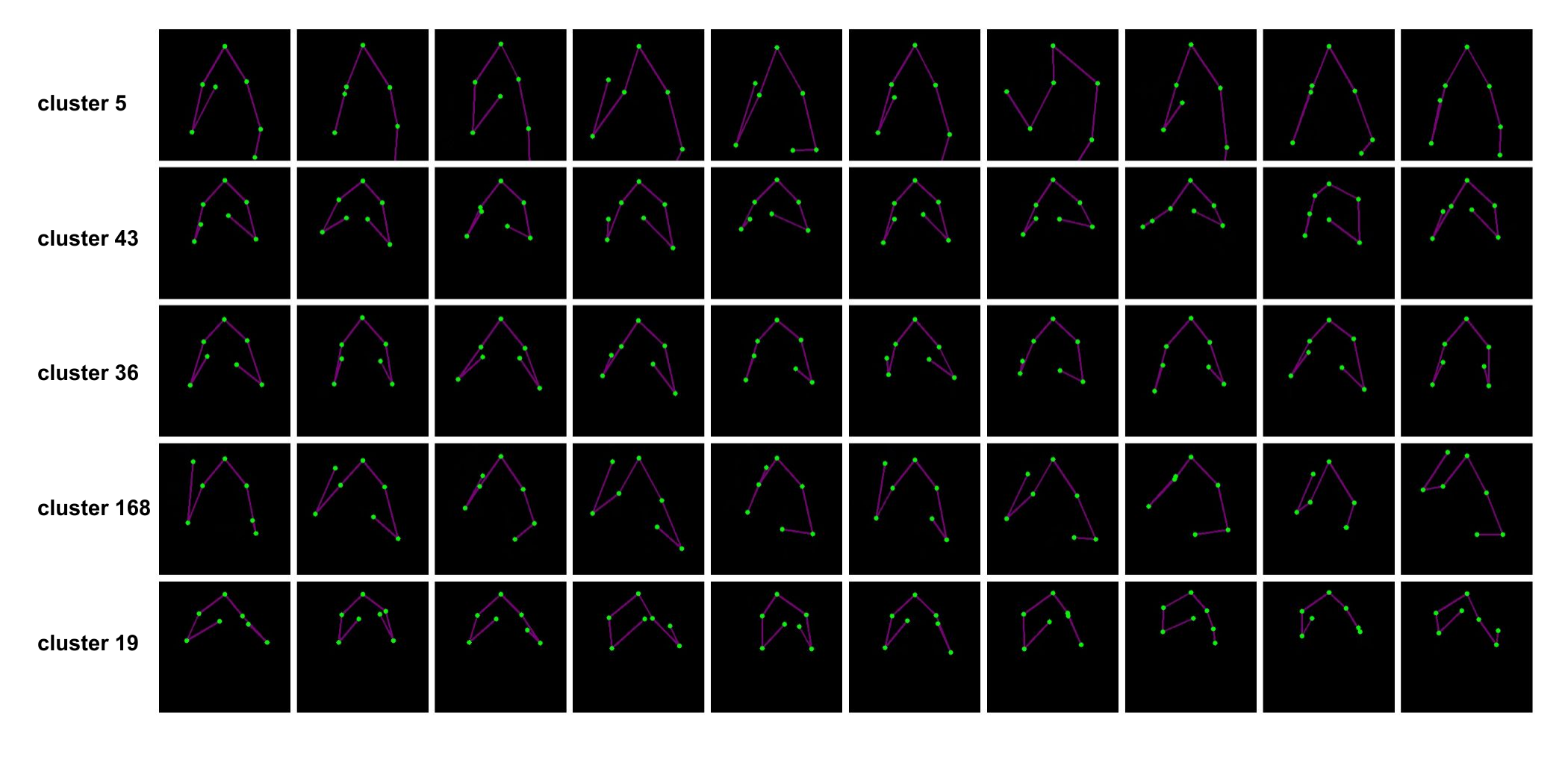}

    \caption{Sample upper body clusters.
    Each row represents a distinct cluster and 10 random examples from it. Cluster 5 seems to correspond to configurations of the upper body pose where the right hand is at shoulder level and the left hand is down. Cluster 43 seems to be a configuration where the two hands are raised and close to each other near the chest, and the signer is facing slightly to the right. This might correspond to signs being performed with both hands involved/active. Cluster 36 appears similar, but with the hands slightly farther apart.
    Cluster 19 is generally similar to cluster 43, except that the signer tends to be facing slightly to the left.  
    Finally in cluster 168, the right hand is usually above the shoulder and close to the face/head.}

    \label{fig:cluster_body}
\end{figure*}

We visualize clustering results for the face, left hand, right hand, and upper body pose in \cref{fig:cluster_face,fig:cluster_lhand,fig:cluster_rhand,fig:cluster_body}. All cluster samples were randomly selected (i.e., without manual curation or cherry-picking). Each row represents a cluster.  For each channel (Face, Left hand, Right hand, Upper Body), we include 10 random examples for 5 random clusters.  While there is variability within each cluster, and some clusters contain a large mix of poses, we can also see a great deal of systematic behavior, where the images in a cluster tend to correspond to similar gestures regardless of signer appearance or other visual properties.  The caption for each figure provides our interpretations of some of the clusters.

\section{Translation examples}
\label{sec:translation_examples}

We provide example translations produced by our model given inputs from three ASL datasets: How2Sign (instructional content, \cref{tab:supp+h2s}), FLEURS-ASL (zero-shot setting, \cref{tab:supp+fleursasl}), and OpenASL (general domain with native signers and varying background, \cref{tab:supp+oasl}). %

\begin{table*}
\centering

\vspace{-10pt}
\resizebox{\linewidth}{!}{
\begin{tabular}{p{0.008\textwidth}p{0.2\textwidth}p{0.8\textwidth}}

\hline
(1) & Reference & And that's a great vital point technique for women's self defense. \\
    & \cite{uthus2023youtube} & It's really great for women's self defense. \\
    &\cite{rust-etal-2024-towards} & This is a really great point for women's self defense. \\
    & Ours & If you're a bigger person we're talking about really self defense here. \\
\hline
(2) & Reference & In this clip I'm going to show you how to tape your cables down. \\
    & \cite{uthus2023youtube} & In this clip we're going to show you how to cut a piece of clay. \\
    &\cite{rust-etal-2024-towards} & In this clip I'm going to show you how to clip the cable, the cable. \\
    & Ours & In this clip I'm going to show you how to brand out the cable strings. \\
\hline
(3) & Reference & In this segment we're going to talk about how to load your still for distillation of lavender essential oil. \\
    & \cite{uthus2023youtube} & In this clip we're going to talk about how to feed a set of baiting lizards for a lava field oil. \\
    &\cite{rust-etal-2024-towards} & In this clip we're going to talk about how to feed the trail for draining clean for laborer oil. \\
    & Ours & In this clip we're going to talk about how to take our stick for disinfectant oil. \\
\hline
(4) & Reference & You are dancing, and now you are going to need the veil and you are going to just grab the veil as far as possible. \\
    & \cite{uthus2023youtube} & Their hopping and dancing is now, they're going to need their squat and squat and they're going to be able to move independently. \\
    &\cite{rust-etal-2024-towards} & So that she's going to get her hips up as far as she can, and now she's going to lift her head up as far as possible. \\
    & Ours & Her dancing and now she needs her feather to grab it with her foot as far as possible. \\
\hline
(5) & Reference & But if you have to setup a new campfire, there's two ways to do it in a very low impact; one is with a mound fire, which we should in the campfire segment earlier and the other way to setup a low impact campfire is to have a fire pan, which is just a steel pan like the top of a trash can. \\
    & \cite{uthus2023youtube} & But if you have to set up a new campfire, there are two ways to do a low impact fire, one is a cone fire, which we have to do in the tent earlier, and the other one is to set up a campfire in a fire pan. \\
    &\cite{rust-etal-2024-towards} & But if you have to set up a new campfire, this is one way to do it in a low impact. One is a monk fire. One is a campfire. The other one is to set a campfire in a campfire. That's just a post like the top of the post. \\
    & Ours & But if you have to set a new campfire, there are two ways to do a low impact one is a bond fire, which we should do in your campfire, another one is to set a campfire in a fire pan that is just just set a pan like the top of it pan. \\
\hline
(6) & Reference & So, this is a very important part of the process. \\
    & \cite{uthus2023youtube} & Alright, let's get started. \\
    &\cite{rust-etal-2024-towards} & It's an important part of the process. \\
    & Ours & This is a very important part of the process. \\
\hline
\end{tabular}
}
\caption{Qualitative translation examples from the How2Sign dataset,  comparing SHuBERT-based translations to previous models.}
\label{tab:supp+h2s}
\end{table*}

\begin{table*}
\centering

\vspace{-10pt}
\resizebox{\linewidth}{!}{
\begin{tabular}{p{0.008\textwidth}p{0.15\textwidth}p{0.8\textwidth}}

\hline
(1) & Reference & During the 1980s he worked on shows such as Taxi, Cheers, and The Tracy Ullman Show.\\
    & \cite{tanzer2024fleurs} & In the 1980s, she worked in theaters like taxesi, cheesy, and tracy. \\
    & Ours & In the 1980's, she worked for theaters like Taxi Chers, Tracy Ullman Shaw. \\
\hline
(2) & Reference & The rise of new technologies allows us to see and investigate brain structures and processes never seen before. \\
    & \cite{tanzer2024fleurs} & There is a new technique to detect brains and vision. \\
    & Ours & Increasing new technology that allows people to consider investigating their brain structures and brain structures. \\
\hline
(3) & Reference & The Articles required unanimous consent from all the states before they could be amended and states took the central government so lightly that their representatives were often absent. \\
    & \cite{tanzer2024fleurs} & The law requires all states to agree on a standard and that it is a legal requirement. \\
    & Ours & The state’s agreement requires all standardized agreements to remove the standards of representatives from the state to represent the state’s amendments. \\
\hline
\end{tabular}
}
\caption{Qualitative translation examples from the FLEURS-ASL dataset, comparing SHuBERT-based translation (zero-shot) to a previous approach~\cite{tanzer2024fleurs}.}%
\label{tab:supp+fleursasl}
\end{table*}

\begin{table*}
\centering

\vspace{-10pt}
\resizebox{\linewidth}{!}{
\begin{tabular}{p{0.03\textwidth}p{0.2\textwidth}p{0.8\textwidth}}

\hline
(1) & Reference & thank you \\
    & \cite{shi2022open} & thank you \\
    & Ours & thank you \\
\hline
(2) & Reference & come on \\
    & \cite{shi2022open} & come on \\
    & Ours &  maybe \\
\hline
(3) & Reference & now i’ve come this far and it ’s a different team \\
    & \cite{shi2022open} & how do you feel about it \\
    & Ours & it feels like a crash in the team \\
\hline
(4) & Reference & i was there from the beginning to the end and time went by fast \\
    & \cite{shi2022open} & the students were thrilled by this \\
    & Ours & i just wanted to leave because it went ahead and started \\
\hline
(5) & Reference & i’m here at nad’s 50th wow \\
    & \cite{shi2022open} & the nad has been <unk> for many years \\
    & Ours & well the nad is shocked to have 50 years of dhh \\
\hline
(6) & Reference & i entered the yap 2018 competition and won \\
    & \cite{shi2022open} & the competition was started with ideas \\
    & Ours & i enrolled in that competition in 2018 and then i won \\
\hline
(7) & Reference & you can check out their kickstarter in the link below \\
    & \cite{shi2022open} & you can watch the conversation at lake county \\
    & Ours & you can check out their kickstarter link below \\
\hline
(8) & Reference & that is one thing i found interesting and wanted to share with you today \\
    & \cite{shi2022open} & i also am the president of the jr. nad conference here \\
    & Ours & that's one interesting thing she wanted to share with you \\
\hline
(9) & Reference & those are the different types of bills \\
    & \cite{shi2022open} & schools have switched to teaching students \\
    & Ours & i looked at several different types of interpreting services \\
\hline
(10) & Reference & dry january has picked up in popularity since it began in 2012 \\
    & \cite{shi2022open} & krispy kreme is bringing back its original playstation in 2016 \\
    & Ours & the qury dry january started in 2012 \\
\hline
(11) & Reference & we will be happy to respond give you support and listen to your concerns \\
    & \cite{shi2022open} & please review and submit your time passion and support this important issue \\
    & Ours & the nad is willing to respond and support your concerns  \\
\hline
(12) & Reference & there were videos posted on the internet that showed a person walking on the grass completely engulfed in flames \\
    & \cite{shi2022open} & a video shows the officer walking up to his shoulder and before he was shot \\
    & Ours &  videos posted on social media of him walking on a grass walking completely with fire \\
\hline
(13) & Reference & and people would become carpenters laborers mechanics plowers and farmers \\
    & \cite{shi2022open} & the next year 1880 the nad was established in the first operation 30 of the house in 2015\\
    & Ours & and he was forced to wear a wearing a labover meganic and a financial warper \\
\hline
(14) & Reference & for nad youth programs related information please contact us via facebook at the nad youth programs or email us through \\
    & \cite{shi2022open} & you can contact us through our website where you can check our facebook page online at <unk> \\
    & Ours & if you want to contact the nad youth program you can contact us through our facebook page at the nad youth program or youth program through our website \\
\hline
(15) & Reference & last week suspects gregory mcmichael and his son travis were arrested and charged with felony murder and aggravated assault \\
    & \cite{shi2022open} & last week a black man named <unk> <unk> was arrested and charged with felony murder and aggravated assault\\
    & Ours &  last week two suspects gregory mcmichael and his son travis were arrested and charged with felony murder and wounded by another gravated assaulter \\
\hline

\end{tabular}
}
\caption{Qualitative translation examples from the OpenASL dataset,  comparing SHuBERT-based translations to a previous model~\cite{shi2022open}}.%
\label{tab:supp+oasl}
\end{table*}

\end{document}